\pdfoutput=1
\documentclass[11pt]{article}

\usepackage[]{acl}

\usepackage{times}
\usepackage{latexsym}
\usepackage{tabularx}
\usepackage{booktabs}
\usepackage{graphics}
\usepackage{graphicx}
\usepackage{subfigure}
\usepackage{amsmath}
\usepackage{cleveref}
\usepackage{amssymb}% http://ctan.org/pkg/amssymb
\usepackage{pifont}% http://ctan.org/pkg/pifont
\usepackage{changepage}

\usepackage{csquotes}
\usepackage{lipsum}

\newcommand{\ie}{\textit{i.e.,~}}
\newcommand{\eg}{\textit{e.g.,~}}

\usepackage[T1]{fontenc}
\usepackage[utf8]{inputenc}

\usepackage{microtype}

\usepackage{inconsolata}

\title{\textsc{Cev-LM}: Controlled Edit Vector Language Model \\ for Shaping Natural Language Generations}

\newcommand*{\affaddr}[1]{#1} % No op here. Customize it for different styles.
\newcommand*{\affmark}[1][*]{\textsuperscript{#1}}
\newcommand*{\email}[1]{\texttt{#1}}

\author{%
Samraj Moorjani\affmark[1], Adit Krishnan\affmark[1], Hari Sundaram\affmark[1] \\
\affaddr{\affmark[1]University of Illinois at Urbana-Champaign, USA}\\
\email{\{samrajm2, aditk2, hs1\}@illinois.edu}\\
}

\begin{document}
\maketitle

% Version 1 (ARR August)
% \begin{abstract}
%     As large-scale language models become the standard for text generation, there is a greater need to tailor the generations to be more or less concise, targeted, and informative, depending on the application. For instance, complex ideas can be presented concisely to an expert, but non-technical audiences may need more context and a slower-paced introduction to grasp the same idea. Current controllable text generation (CTG) systems can incorporate well-defined control conditions such as keyword inclusion and syntactical structure. However, these controls are insufficient to measure and modulate complex objectives such as speed which require controlling the shape of the text. In this paper, we introduce \textsc{Cev-LM} - a lightweight, semi-autoregressive language model that utilizes constrained edit vectors to control three complementary metrics (speed, volume, and circuitousness) that quantify the shape of text (\eg pacing of content). We study an extensive set of state-of-the-art CTG models and find that \textsc{Cev-LM} provides significantly more targeted and precise control of these three metrics while using less training data and fewer parameters. Thus, we can shape the presentation of the same semantic content to suit different audiences for improved readability. Our code and data are accessible at \url{https://drive.google.com/file/d/10rwCLJ96eNP5LS_1sG-flWvXD9X4pbjO}.
    
%     % \textsc{Cev-LM} relies on two novel modifications to an edit-then-prototype model: constrained neighborhood creation and controlled edit vector perturbation. 
    
% \end{abstract}

\begin{abstract}
    As large-scale language models become the standard for text generation, there is a greater need to tailor the generations to be more or less concise, targeted, and informative, depending on the audience/application. Existing control approaches primarily adjust the semantic (\eg emotion, topics), structural (\eg syntax tree, parts-of-speech), and lexical (\eg keyword/phrase inclusion) properties of text, but are insufficient to accomplish complex objectives such as pacing which control the complexity and readability of the text. In this paper, we introduce \textsc{Cev-LM} - a lightweight, semi-autoregressive language model that utilizes constrained edit vectors to control three complementary metrics (speed, volume, and circuitousness) that quantify the shape of text (\eg pacing of content). We study an extensive set of state-of-the-art CTG models and find that \textsc{Cev-LM} provides significantly more targeted and precise control of these three metrics while preserving semantic content, using less training data, and containing fewer parameters.\footnote{Our code and data are accessible at this link \url{https://github.com/CrowdDynamicsLab/CEVLM}}
\end{abstract}

\section{Introduction}
\label{sec:introduction}

\begin{figure}[!htb]
  \centering
  \caption{Generated examples of change in speed, volume, and circuitousness, metrics that define the shape of text, and stylized illustrations. The points represent the word embeddings of windows of text, $\{x_1, ..., x_n\}$. The original text has a lower value of the metric, and our generation ($\textsc{Cev-LM}$) demonstrates a higher value.  }
  \vspace{-0.1cm}
  \scalebox{0.65}{
    \includegraphics[width=0.7\textwidth]{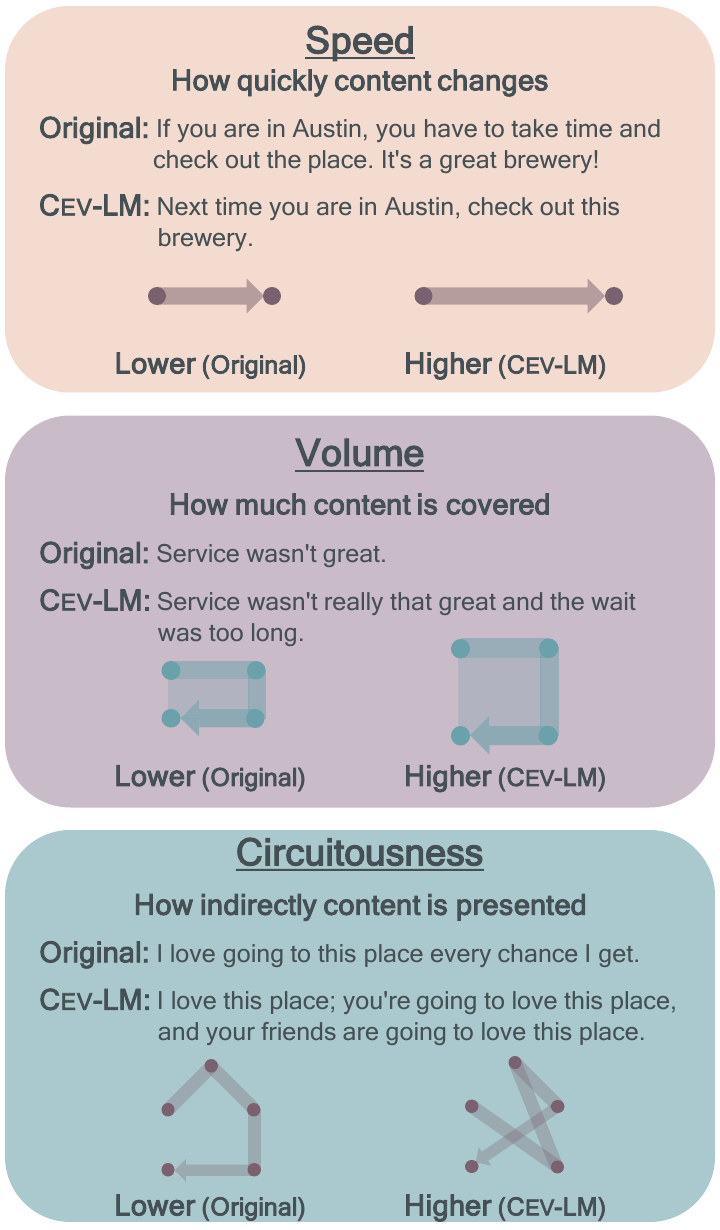}
  }
  \vspace{-0.5cm}
  \label{fig:front_page_ex}
\end{figure}

As large-scale pre-trained language models allow the generation of more diverse and fluent text, controllable text generation (CTG) is crucial to meet the needs of different applications and audiences. For instance, complex ideas can be presented concisely to an expert, but non-technical audiences may need more context and a slower-paced introduction to grasp the same idea. Existing CTG approaches empirically evaluate three types of control conditions: semantic (\eg emotion, topics), structural (\eg syntax tree, parts-of-speech), and lexical controls (\eg keyword/phrase inclusion) \cite{Zhang2022ASO}. While this taxonomy covers a broad range of features, it does not target more complex objectives, such as the pacing of text. \citet{toubia-2021} presents a set of measures that quantify the shape of narratives, relying on both semantic and structural properties of the text. Speed measures how quickly content changes, volume quantifies how much content is covered, and circuitousness represents how indirectly content is presented.  
% \citet{toubia-2021} presents a set of measures (\ie speed, volume, and circuitousness) that quantify the shape of narratives, relying on both semantic and structural properties of the text.
% For instance, while some generated texts can be targeted and concise, others seem outstretched and redundant.

Controlling these ``nonstandard'' control conditions, such as speed, volume, and circuitousness, is challenging because they are built on interconnected semantic, structural, and lexical properties. CTG approaches have been developed and tested separately for semantic, structural, or lexical features, but not at the intersection of multiple features \cite{Zhang2022ASO, li2022diffusion}. Furthermore, these nonstandard control conditions require sentence and paragraph-level reconstruction. This is challenging for purely autoregressive approaches, which struggle with longer context lengths \cite{beltagy2020longformer}. Conversely, deep generative model-based approaches, such as \citet{guu2018generating}, produce generations from a continuous latent variable, enabling simple, gradient-based methods to perform complex control tasks over larger contexts \cite{li2022diffusion, han2022ssd}.  

In summary, our contributions are as follows: firstly, we present the $\textsc{Cev-LM}$ framework to provide a lightweight, ``tuning-knob'' to control speed, volume, and circuitousness. We adopt a semi-autoregressive paradigm to exploit both the generation quality of autoregressive models and the controllability of deep generative models. Second, we propose a controlled edit vector approach where we preselect examples from a constrained similarity neighborhood to match our criteria and apply a controlled edit vector in the latent space to tune the desired attribute. Lastly, we study a robust set of benchmarks used in CTG and demonstrate that \textsc{Cev-LM} provides significantly more control and preserves both relevance and fluency (\S\ref{sec:ne_results}) across both high and low-resource settings while using fewer training samples and parameters.

\section{Related Work}
\label{sec:rel_works}

In this section, we briefly introduce existing literature on controllable text generation (\S\ref{rel:ctg}) and the prototype-then-edit architecture (\S\ref{rel:prototype}).

\subsection{Controllable Text Generation}
\label{rel:ctg}

\citet{Zhang2022ASO} find that existing controllable text generation (CTG) methods fall under three major categories: fine-tuning \cite{li2021prefix, Tambwekar_2019, ouyang2022training}, retraining/refactoring \cite{keskar2019ctrl}, and post-processing \cite{Dathathri2020Plug, scialom2020discriminative, krause2020gedi, kumar2021controlled}. \citet{li2021prefix} train a small, continuous, task-specific vector prepended to the input of a pretrained language model (PLM), keeping the parameters of the PLM frozen, providing a lightweight alternative to fine-tuning. \citet{krause2020gedi} guides the generation of a larger PLM using two class-conditional language models, one conditioned on the desired control and one conditioned on the ``anti-control''. \citet{kumar2021controlled} replaces traditional decoding with a continuous optimization problem, where desired controls can be expressed as multiple differentiable constraints. 

Many CTG works utilize deep generative models such as variational auto-encoders (VAEs) \cite{guu2018generating, xu2020variational, wang2019topic}, generative adversarial networks (GANs) \cite{scialom2020discriminative}, and diffusion models \cite{li2022diffusion, han2022ssd} because of the malleability of the latent state. However, recent work has relied on plug-and-play approaches with large-scale pretrained language models (PLMs) without significant task-specific retraining. The autoregressive design of PLMs makes it challenging to exhibit control on sentence- and paragraph-level constraints such as speed, volume, and circuitousness \cite{toubia-2021}. Further, despite the benefits of fine-tuning and post-processing-based approaches, more direct control is necessary \cite{soatto2023taming}. While discretized controls are more natural (\eg less vs. more toxic), we emphasize a ``tuning-knob''-like control as it provides more fine-grained control, and it is trivial to go from continuous to discrete controls, but not the converse.

\subsection{Prototype-then-Edit}
\label{rel:prototype}

Prototype editing applies attribute markers to predefined sentence templates to generate sentences that are semantically similar but altered content~\cite{guu2018generating, li2018delete, sudhakar-etal-2019-transforming}. \citet{guu2018generating} introduce an unconditional generative model that samples a ``prototype'' sentence from the training corpus and edits it using a randomly sampled edit vector. In the Yelp restaurant review corpus, 70\% of the test set is within a Jaccard distance of 0.5 of a training set sentence, implying that a neural editor with smooth and consistent edits should capture the test set. The edit model has two significant constraints: semantic smoothness and consistent edit behavior. Specifically, edits should change the semantics of text by a small amount and, when stacked together, create a more significant change. Further, the edit vector, $z$, should control the change in a sentence such that when applied to different sentences, the edits are semantically analogous. We adopt the prototype-then-edit framework because of the ability of the edit vector to reflect a desired change in attribute in the latent space, hence ``controlled edit vectors''.

\section{Nonstandard Control Conditions}
\label{sec:attributes}

We pick three non-standard control metrics to evaluate our approach: speed, volume, and circuitousness \cite{toubia-2021}. Speed is a measure of how quickly content moves in a given text,  calculated as the distance traveled by consecutive windows of text. Specifically, speed is equal to $\frac{\sum_{t=1}^{T-1} \|x_{t+1} - x_t\|}{T-1}$ where $x_t$ is the word embedding of the $t-th$ window of text. Volume captures the amount of information covered in a piece of text, calculated by finding the minimum volume ellipsoid that contains all $x_t, \forall t \in \{1 \dots T\}$. Circuitousness measures how \textit{indirectly} content is covered and is formulated as $\frac{\sum_{t=1}^{T-1} \|x_{t+1} - x_t\|}{L_{SP}}$ where $L_{SP}$ is the length of the shortest path, computed with the traveling salesman problem. While volume measures how much content is covered, circuitousness answers how that content was covered. Given $s(\cdot)$ to compute the target attribute, we define the control of generated text as how close $s(x) - s(x')$ is to a desired change in attribute, $\Delta$, where $x$ and $x'$ are the generated and original text, respectively. These measures have been used to study the success of narratives and can be used to quantify complex control objectives, such as how concise or informative generations are.

\section{Controllable Edit Vectors}
\label{sec:ne_edits}

% Note that  
The prototype-then-edit architecture \cite{guu2018generating} features three main components: a neural editor $p_{edit}(x | x', z)$, an inverse neural editor $q(z | x', x)$, and an edit prior $p(z)$. The inverse neural editor and neural editor combine to form the encoder and decoder of a variational autoencoder \cite{kingma2013auto}, respectively. The neural editor is implemented as an autoregressive, sequence-to-sequence model with attention, where given $x'$ as input and $z$, which is concatenated to the input of the decoder at each step, the model generates $x$. The edit prior is defined as $z = z_{norm} \cdot z_{dir}$ where $z_{norm}$, the strength of the edit, is drawn from $\mathcal{U}(0,10)$ and $z_{dir}$, the direction of the edit, is sampled from a uniform distribution on the unit sphere. Note that both $z_{dir}$ and $z_{norm}$ are vectors. The inverse neural editor is given the edit pair $(x, x')$ and must infer the edit vector $z$. The difference between $x$ and $x'$ is represented as

$$f(x, x') = \sum_{w \in I} \Phi(w) \oplus \sum_{w \in D} \Phi(w)$$

where $I = x \setminus x'$ (\ie the set of words added to $x'$), $D = x \setminus x'$ (\ie the set of words deleted from $x'$), $\Phi(w)$ is the GloVe \cite{pennington2014glove} vector for $w$, and $\oplus$ is the concatenation operation. The inverse neural editor infers the edit vector through a perturbed version of $f(x, x')$ , as follows: \vspace{-5pt}

$$q(z_{dir} | x', x) = \text{vMF}(f_{dir}, \kappa)$$
$$q(z_{norm} | x', x) = \mathcal{U}(f_{norm}, f_{norm} + \epsilon)$$

where $f_{norm} = \min(\|f\|, 10 - \epsilon)$ and $f_{dir} = \frac{f}{f_{norm}}$. Let $\text{vMF}(\mu, \kappa)$ be a von-Mises Fisher distribution where $\mu$ is the mean vector, and $\kappa$ is the concentration parameter, controlling the decay rate.

To exhibit control over our target attributes, we alter the prototype-then-edit model in two ways: neighborhood creation and edit vector perturbation. The former constrains the inferred edit vector to demonstrate the desired change in attribute within some tolerance $\epsilon$. The latter encourages a perturbation to the edit vector in the desired direction to compensate for $\epsilon$.

\textbf{Constrained Neighborhood Creation}: The likelihood of a sentence is formulated as $p(x) = \sum_{x' \in \mathcal{X}} p(x | x') p(x')$ where $x'$ is prototype sentence and $x$ is the generated sentence. The likelihood $p(x | x')$ is defined as $\mathbb{E}_{z \sim p(z)} [p_{edit}(x | x', z)]$. A sum over all prototypes $x'$ is expensive, so we only sum over the $x'$ that are lexically similar to $x$ - a lexical similarity neighborhood, $\mathcal{N}(x)$. Further, we create an additional constraint on the target attribute to ensure that inferred edit vectors from the inverse neural editor correspond to a specified change in that attribute. More formally, we define the neighborhood with a tolerance $\epsilon$ as \vspace{-15pt}

\begin{align*}
    \mathcal{N}_{\Delta}(x) &= \{x' \in \mathcal{X} : d_J(x, x') < 0.5, \\
                            &  \hspace{0.3in} |(s(x) - s(x')) - \Delta| \leq \epsilon \}
\end{align*}

\textbf{Controlled Edit Vector Peturbation}: We hypothesize that by altering the magnitude of $z_{norm}$ and the direction $z_{dir}$, we can control the strength and behavior of the edit vector. Expressly, we can condition the formulation of the inverse neural editor on the target attribute by defining $q(z_{norm} | x', x) = \mathcal{N}(\Delta, 1) \cdot \mathcal{U}(f_{norm}, f_{norm} + \epsilon)$, where $\mathcal{N}$ is the normal distribution and $\mathcal{U}$ is the uniform distribution.

% In order to introduce a control mechanism for our target features, we alter two aspects of the prototype-then-edit model: neighborhood creation and edit vector perturbation.

% Note that direction is enforced with $(s(x) - s(x'))$.

\section{Experimental Settings}
\label{sec:ne_exp_settings}

We train variants of \textsc{Cev-LM} on the Yelp Restaurant Reviews Corpus~\cite{yelp2017}. The corpus has over 5.84 million training and 2.08 million test reviews (English) \footnote{We limit the test set to 1000 samples due to the cost of the OpenAI API.} in the original similarity neighborhood (\ie Jaccard distance less than $0.5$\footnote{The Jaccard distance is tuned by~\cite{guu2018generating}.}). The dataset provides a broad variety of writing styles (\eg formal vs. informal, positive vs. negative sentiment) and topics (\eg hotels, food, service, etc.) to test our approach~\cite{gong2017clustered, guu2018generating, chu2019meansum}.

% \textit{\textbf{Baseline}} We use the out-of-the-box model to provide a baseline on sequence-to-sequence evaluation metrics as well as a baseline $\Delta$.

\vspace{0.1in}\textbf{\textsc{Cev-LM ($\mathcal{N}$-only)}:} We add the $\Delta$ constraint during neighborhood creation and train on the newly formed data. % We try $\epsilon = 0.05, 0.1, 0.2$ as tolerance values.

\vspace{0.1in}\textbf{\textsc{Cev-LM}:} We use both modifications, constrained neighborhood creation, and controlled edit vector perturbation. % testing $\epsilon = 0.05, 0.1, 0.2$.

We provide the hyperparameters for our experiments in~\Cref{appendix:cevlm_params}. As our method generally falls under retraining \cite{Zhang2022ASO}, we provide an extensive set of benchmarks for the fine-tuning and post-processing categories, relying on both deep generative models (\ie diffusion) and autoregressive architectures.

% We compare our approach against multiple state-of-the-art baselines: % in controllable text generation: 

\vspace{0.1in}\textbf{GPT-3:} We construct multiple few-shot prompts for all attributes to generate a sentence with the desired change in attribute given a sentence. The prompts consist of three parts: a language-based description of the attribute, $n$ examples of the desired change in attribute, and the prompt to generate a sentence. We use the ``davinci'' model for all experiments and describe the process further in~\Cref{appendix:gpt}.

\vspace{0.1in}\textbf{MuCoCO:} \textbf{Mu}ltiple \textbf{Co}nstraints through \textbf{C}ontinuous \textbf{O}ptimization \cite{kumar2021controlled} is an alternative to fine-tuning for controllable text generation that formulates decoding as a continuous optimization problem with multiple differentiable constraints. To control the three attributes, we define a constraint $|(s(x) - s(y)) - \Delta|$ where $x$ and $y$ are the input and output sentences. We train a regressor, $\mathcal{D}(x,y)$, to approximate $s(x) - s(y)$ because the computation of speed, volume, and circuitousness is not differentiable.  We present the mean absolute error (MAE) and normalized mean absolute error (NMAE) in~\Cref{tab:clf_reg_eval}. The MAEs are relatively small compared to the scale of $\Delta$, reflected in the NMAE, indicating a strong regressor. We provide more details on the training of the regressor in~\Cref{appendix:crt}.

% \citet{kumar2021controlled} formulate decoding as a continuous optimization process with differentiable constraints to allow for controllable natural language generation. We train robust regressor for each feature and utilize MuCoCO to generate sentences with a desired change in feature. We provide more details in~\Cref{appendix:mucoco}.

\begin{table}[]
\centering
  \caption{Evaluation metrics for the regressor, $\mathcal{D}(x,y)$, and classifier model, $\mathcal{C}(x,y)$, used in MuCoCO and SSD-LM across all control attributes. Note that $x$ and $y$ are two sentences, and the goal is to predict the difference in attribute, either directly or within a bin. For $\mathcal{D}$, we report mean absolute error (MAE) and a normalized mean absolute error (NMAE) and for $\mathcal{C}$, we report F1, MAE, and NMAE.}
  \label{tab:clf_reg_eval}
\scalebox{0.65}{
\begin{tabularx}{1.5\linewidth}{cccccc}
    \toprule[1.5pt]
                   & $\mathcal{D}$\textsc{-MAE} & $\mathcal{D}$\textsc{-NMAE} & $\mathcal{C}$\textsc{-F1} & $\mathcal{C}$\textsc{-MAE} & $\mathcal{C}$\textsc{-NMAE} \\
    \midrule[0.75pt]
    Speed          & 0.3013 & 0.0764 & 0.6533 & 0.4979 & 0.1263 \\
    Volume         & 0.2144 & 0.1034 & 0.5922 & 0.6196 & 0.2987 \\
    Circuitousness & 0.0459 & 0.0236 & 0.6763 & 0.3730 & 0.1917 \\
    \bottomrule[1.5pt]\\   
\end{tabularx}
}
\vspace{-15pt}
\end{table}

\newcommand*{\newfactor}{0.056}
\begin{table*}[!htbp]
\centering
\tiny
\caption{Achieved delta for speed, volume, and circuitousness across all approaches for different target deltas. The scores are averaged across three training runs (inference runs for GPT-3). We use a tolerance $\epsilon = 0.1$ for all of our approaches, as it empirically provided the best results in~\Cref{subsec:tolerance}. We find that our approaches (\textsc{Cev-LM ($\mathcal{N}$-only)} and \textsc{Cev-LM}) show significantly more control over all control conditions across nearly all target deltas.\vspace{-5pt}}
%\footnotesize
\noindent\setlength\tabcolsep{2.9pt}
% \fontsize{8pt}{7pt}\selectfont
\resizebox{\textwidth}{!}{%
\begin{tabular}{@{}p{0.20\linewidth}@{\hspace{2pt}}
% K{\newfactor\linewidth}K{\newfactor\linewidth}K{\newfactor\linewidth}@{\hspace{5pt}}
% K{\newfactor\linewidth}K{\newfactor\linewidth}K{\newfactor\linewidth}@{\hspace{5pt}}
% K{\newfactor\linewidth}K{\newfactor\linewidth}K{\newfactor\linewidth}@{\hspace{5pt}}
% K{\newfactor\linewidth}K{\newfactor\linewidth}K{\newfactor\linewidth}@{}} \\
ccc@{\hspace{5pt}}ccc@{\hspace{5pt}}ccc@{\hspace{5pt}}ccc@{}} \\
\toprule
{\textbf{Metric}} & \multicolumn{4}{c}{\textbf{Speed}}  & \multicolumn{3}{c}{\textbf{Volume}} & \multicolumn{3}{c}{\textbf{Circuitousness}} & \multicolumn{1}{c}{\textbf{\% Err}}\\
\cmidrule{2-5} \cmidrule(lr){6-8} \cmidrule(lr){9-11} \cmidrule{12-12}
%\midrule
 {\textbf{Target Delta}} & \textbf{0.125} & \textbf{0.5} & \textbf{2.0} & \textbf{4.0} & \textbf{0.125} & \textbf{0.5}& \textbf{2.0} & \textbf{0.125} & \textbf{0.5}& \textbf{1.0} & -\\

\midrule
\multicolumn{11}{c}{\textsc{Benchmark Approaches}} \\
\midrule[0pt]

\textbf{GPT-3}~\cite{brown2020language}       & -0.091 & 0.023  & 0.200 & 0.294 & 0.019 & 0.150 & 1.769 & 0.011  & 0.009         & 0.021 & 90.44 \\
\textbf{MuCoCO}~\cite{kumar2021controlled}   & 0.169 & 0.871 & 1.293 & 3.610 & 0.142 & 0.558 & 2.310 & 0.068  & 0.058         & 0.054 & 42.38 \\
\textbf{SSD-LM}~\cite{han2022ssd}    & 0.412 & 0.335  & 1.094 & 1.059 & 0.369 & 0.283 & 0.671 & \textbf{0.075}  & 0.078         & 0.109 & 90.00 \\
\textbf{Prefix Tuning}~\cite{li2021prefix}  & 0.197 & 0.245 & 1.562 & -       & 0.088 & 0.877 &  -      & -0.004 & 0.159         & -  & 58.12 \\

\midrule[0pt]
\multicolumn{11}{c}{\textsc{Our Approaches}} \\
\midrule[0pt]

\textbf{\textsc{Cev-LM} ($\mathcal{N}$-only)} & 0.111 & \textbf{0.457} & \textbf{1.760} & \textbf{3.621} & 0.111 & 0.443 & 1.752  & \textbf{0.072}  & 0.423          & \textbf{0.790} & 15.51 \\
\textbf{\textsc{Cev-LM}} & \textbf{0.118} & 0.450 & 1.755 & 3.547 & \textbf{0.114} & \textbf{0.451} & \textbf{1.863} & 0.064  & \textbf{0.431}         & 0.781 & \textbf{14.91}\\

\bottomrule
\end{tabular}}
\vspace{-10pt}
\label{tab:errors}
\end{table*}

\vspace{0.1in}\textbf{SSD-LM:} Semi-autoregressive Simplex-based Diffusion Language Model (SSD-LM) \cite{han2022ssd} utilizes diffusion-based language modeling in an iterative manner to generate flexible length text. The diffusion is performed on the vocabulary space allowing for classifier feedback and hence controllable generation. Continuous diffusion models are formulated well for modular control by utilizing gradients from an auxiliary model (\eg use a sentiment classifier to guide the output of a language model to have positive sentiment). We train a classifier to predict a binned difference in attributes such that all bins contain an equal number of training samples. We record both F1-score as well as the MAE and NMAE between classes in~\Cref{tab:clf_reg_eval}. The class labels are generally off by at most one due as shown by the low MAEs, indicating a strong classifier. In~\Cref{appendix:crt}, we describe how we train the classifier model, $\mathcal{C}(x,y)$, to guide generations.

\vspace{0.1in}\textbf{Prefix Tuning:} \citet{li2021prefix} propose prefix-tuning, a lightweight, modular alternative to fine-tuning that trains a small continuous vector prepended to the input (\ie a prefix) while keeping the parameters of the language model frozen. The approach is similar to prompt-tuning but allows the task-specific prefix to consist entirely of free parameters. We use the same settings as the abstractive summarization experiment in the original paper, using BART \cite{lewis2019bart} with a prefix sequence length of $200$. We freeze the aforementioned regressor and add $\gamma |(s(x) - s(y)) - \Delta|$ to the existing loss, where $\gamma$ is a tunable parameter. We found that $\gamma = 0.1$ yielded the best results.
% \citet{han2022ssd} present Semi-autoregressive Simplex-based Diffusion Language Model (SSD-LM) which allows for diffusion-based generation with flexible output length during decoding. The diffusion is performed on the vocabulary space allowing for classifier feedback and hence controllable generation. We train a robust classifier for each feature and generate sentence with a desired change in feature. See~\Cref{appendix:ssdlm} for more details.

\vspace{0.1in}We record three main evaluation metrics to test the strength of the control and ensure generations are relevant. \textbf{Delta} measures the change in attribute (\ie $\Delta$). We report the difference as percent error. \textbf{BiLingual Evaluation Understudy (BLEU)} \cite{Papineni2002bleu} measures the n-gram overlap (lexical similarity) and \textbf{BERTScore} \cite{zhang2019bertscore} measures the semantic similarity to ensure that generations remain on topic. We compute the BLEU and BERTScore between the generated and original sentence to ensure the generations remain lexically and semantically similar to the original content.

\section{Results}
\label{sec:ne_results}

% PERCENT ERROR COMPUTATION SCRIPT
% s = s.replace("&", "").replace("\textbf{", "").replace("}", "").strip()
% while '  ' in s:
%     s = s.replace('  ', ' ')
    
% l = s.split()
% r = [0.125, 0.5, 2.0, 4.0, 0.125, 0.5, 2.0, 0.125, 0.5, 1.0]

% p = 0
% c = 0
% for x, y in zip(l, r):
%     if x == "-":
%         c += 1
%         continue
%     p += abs((float(x) - y)/y)
    
% print(p / (len(l) - c))

In this section, we evaluate \textsc{Cev-LM} on the strength of control (\Cref{subsec:control_eval}) as well as the relevance to the original text (\Cref{subsec:semantic_sim} and~\Cref{subsec:lexical_sim}). In \Cref{subsec:tolerance}, we discuss tuning the tolerance hyperparameter and in~\Cref{subsec:tolerance}, the effect of the data distribution on training. We provide qualitative results in~\Cref{subsec:qualitative}.

\subsection{Control Evaluation}
\label{subsec:control_eval}

\Cref{tab:errors} show the achieved delta of each approach across various target deltas along with the average percent error, while~\Cref{tab:bleu_bert} shows the BERT and BLEU scores. We run all baselines three times and report the average. We find that \textsc{Cev-LM} exhibits significantly greater control of $\Delta$ over the baselines while preserving lexical and semantic similarity across all three attributes and all target deltas.

The baselines generally yield fluent but not controlled text. GPT-3 often generates texts with minimal change in attribute (\ie $\Delta = 0$), showing it cannot understand these nonstandard control conditions through few-shot learning. Conversely, prefix-tuning can replicate the attributes somewhat well but falls short due to neural hallucinations and poor-quality text. In low-resource scenarios (\ie high target deltas and fewer training samples), prefix-tuning led to significant over- or under-fitting; thus, we omit the results. In many cases, MuCoCO and SSD-LM perform poorly in terms of percent error but sometimes outperform or perform on par with our approaches. While we cannot fully explain this behavior, we hypothesize that the data distribution of $\Delta$ in $\mathcal{N}(x)$ plays a significant role. In some cases, we see strong results with both \textsc{Cev-LM} modifications, indicating that combining the modifications is beneficial with certain attributes and when $\epsilon$ is tuned. Specifically, volume consistently benefits from controlled edit vector perturbation, while speed and circuitousness show conflicting results. We find that circuitousness has much larger errors on average, likely due to the dependency on computing the shortest path.

We also evaluate on more commonly studied control attributes, formality and toxicity, and present the results in~\Cref{tab:other_attr_error}. Due to its definition, it is likely that \textsc{Cev-LM} is more suited to handle attributes defined with word embeddings. This seems to be reflected in the higher overall percent error, but \textsc{Cev-LM} still produces more controlled generations on both attributes than all other baselines, indicating the robustness of our approach.

\begin{table*}[!htbp]
\centering
\tiny
\caption{Achieved delta for toxicity and formality across all approaches for different target deltas. The scores are averaged across three training runs (inference runs for GPT-3). We use a tolerance $\epsilon = 0.1$ for all of our approaches, as it empirically provided the best results in~\Cref{subsec:tolerance}. We find that our approaches (\textsc{Cev-LM ($\mathcal{N}$-only)} and \textsc{Cev-LM}) show significantly more control over all control conditions across nearly all target deltas.\vspace{-5pt}}
%\footnotesize
\noindent\setlength\tabcolsep{2.9pt}
\fontsize{5pt}{6pt}\selectfont
\resizebox{\textwidth}{!}{%
\begin{tabular}{@{}p{0.20\linewidth}@{\hspace{2pt}}
% K{\newfactor\linewidth}K{\newfactor\linewidth}K{\newfactor\linewidth}@{\hspace{5pt}}
% K{\newfactor\linewidth}K{\newfactor\linewidth}K{\newfactor\linewidth}@{\hspace{5pt}}
% % K{\newfactor\linewidth}K{\newfactor\linewidth}K{\newfactor\linewidth}@{\hspace{5pt}}
% K{\newfactor\linewidth}K{\newfactor\linewidth}K{\newfactor\linewidth}@{}} \\
ccc@{\hspace{5pt}}ccc@{\hspace{5pt}}ccc@{}}
\toprule
{\textbf{Metric}} & \multicolumn{3}{c}{\textbf{Toxicity}}  & \multicolumn{3}{c}{\textbf{Formality}} & \multicolumn{1}{c}{\textbf{\% Err}}\\
\cmidrule{2-4} \cmidrule(lr){5-7} \cmidrule{8-8}
%\midrule
{\textbf{Target Delta}} & \textbf{0.1} & \textbf{0.5} & \textbf{0.9} & \textbf{0.1} & \textbf{0.5} & \textbf{0.9}& - \\

\midrule
% Metric        &       & Toxicity &       &       & Formality &       & \multicolumn{1}{c}{% Err} \\
% Target Delta  & 0.1   & 0.5      & 0.9   & 0.1   & 0.5    & 0.9   & \multicolumn{1}{c}{-}     \\ \hline
% GPT-3         & 0.017 & 0.083    & 0.360 & 0.279 & 0.309  & 0.370 & 83.74                     \\
% MuCoCO        & 0.014 & 0.041    & 0.063 & 0.234 & 0.148  & 0.155 & 92.99                     \\
% SSD-LM        & 0.011 & 0.218    & 0.342 & 0.008 & 0.376  & 0.650 & 58.66                     \\
% Prefix Tuning & 0.188 & 0.397    & 0.978 & 0.273 & 0.339  & 0.941 & 54.50                     \\ \hline
% CEV-LM        & 0.075 & 0.325    & 0.709 & 0.162 & 0.427  & 0.810 & 27.97                    
\textbf{GPT-3}~\cite{brown2020language}      & 0.017 & 0.083    & 0.360 & 0.279 & 0.309  & 0.370 & 83.74 \\
\textbf{MuCoCO}~\cite{kumar2021controlled}   & 0.014 & 0.041    & 0.063 & 0.234 & 0.148  & 0.155 & 92.99 \\
\textbf{SSD-LM}~\cite{han2022ssd}            & 0.011 & 0.218    & 0.342 & 0.008 & 0.376  & 0.650 & 58.66  \\
\textbf{Prefix Tuning}~\cite{li2021prefix}   & 0.188 & \textbf{0.397}    & 0.978 & 0.273 & 0.339  & \textbf{0.941} & 54.50 \\
\midrule[0pt]
\textbf{\textsc{Cev-LM} (Ours)}                     & \textbf{0.075} & 0.325    & \textbf{0.709} & \textbf{0.162} & \textbf{0.427}  & 0.810 & \textbf{27.97}\\

\bottomrule
\end{tabular}}
% \vspace{-10pt}
\label{tab:other_attr_error}
\end{table*}

\subsection{Semantic Similarity}
\label{subsec:semantic_sim}

\begin{table*}[h]
\centering
\tiny
\caption{BLEU and BERT (F1) Scores for speed, volume, and circuitousness across all approaches for different target deltas. The scores are averaged across three training runs (inference runs for GPT-3). We use a tolerance $\epsilon = 0.1$ for all of our approaches, as it empirically provided the best results in~\Cref{subsec:tolerance}.\vspace{-5pt}}
%\footnotesize
\noindent\setlength\tabcolsep{2.9pt}
% \fontsize{9pt}{9pt}\selectfont
\resizebox{\textwidth}{!}{%
\begin{tabular}{@{}p{0.20\linewidth}@{\hspace{2pt}}
% K{\newfactor\linewidth}K{\newfactor\linewidth}K{\newfactor\linewidth}@{\hspace{5pt}}
% K{\newfactor\linewidth}K{\newfactor\linewidth}K{\newfactor\linewidth}@{\hspace{5pt}}
% K{\newfactor\linewidth}K{\newfactor\linewidth}K{\newfactor\linewidth}@{\hspace{5pt}}
% K{\newfactor\linewidth}K{\newfactor\linewidth}K{\newfactor\linewidth}@{}} \\
ccc@{\hspace{5pt}}ccc@{\hspace{5pt}}ccc@{\hspace{5pt}}ccc@{}}
\toprule
{\textbf{Metric}} & \multicolumn{4}{c}{\textbf{Speed}}  & \multicolumn{3}{c}{\textbf{Volume}} & \multicolumn{3}{c}{\textbf{Circuitousness}} \\
\cmidrule{2-5} \cmidrule(lr){6-8} \cmidrule(lr){9-11}
%\midrule
{\textbf{Target Delta}} & \textbf{0.125} & \textbf{0.5} & \textbf{2.0} & \textbf{4.0} & \textbf{0.125} & \textbf{0.5}& \textbf{2.0} & \textbf{0.125} & \textbf{0.5}& \textbf{1.0}\\

\midrule
\multicolumn{11}{c}{\textsc{\textbf{BERTScore} - Benchmark Approaches}} \\
\midrule[0pt]

\textbf{GPT-3}~\cite{brown2020language}       & 0.910 & 0.919 & 0.916 & 0.904 & 0.914 & 0.916 & 0.887 & 0.89 & 0.881 & 0.832 \\
\textbf{MuCoCO}~\cite{kumar2021controlled}   & 0.657 & 0.760 & 0.571 & 0.728 & 0.670 & 0.684 & 0.615 & 0.740 & 0.568 & 0.552 \\
\textbf{SSD-LM}~\cite{han2022ssd}    & 0.841 & 0.840 & 0.823 & 0.821 & 0.835 & 0.833 & 0.825 & 0.846 & 0.840 & 0.822  \\
\textbf{Prefix Tuning}~\cite{li2021prefix}  & 0.895 & 0.904 & 0.891 & - & 0.879 & 0.901 & - & 0.888 & 0.850 & - \\

\midrule[0pt]
\multicolumn{11}{c}{\textsc{\textbf{BERTScore} - Our Approaches}} \\
\midrule[0pt]

\textbf{\textsc{Cev-LM} ($\mathcal{N}$-only)} & \textbf{0.935} & 0.934 & \textbf{0.929} & 0.919 & \textbf{0.938} & \textbf{0.932} & \textbf{0.925} & 0.927 & \textbf{0.911} & \textbf{0.909} \\
\textbf{\textsc{Cev-LM}} & \textbf{0.935} & \textbf{0.939} & \textbf{0.928} & \textbf{0.923} & 0.935 & \textbf{0.931} & 0.921 & \textbf{0.931} & \textbf{0.909} & 0.835 \\

\midrule
\multicolumn{11}{c}{\textsc{\textbf{BLEU} - Benchmark Approaches}} \\
\midrule[0pt]

\textbf{GPT-3}~\cite{brown2020language}       & 0.233 & 0.261 & \textbf{0.321} & 0.225 & 0.219 & \textbf{0.304} & 0.203 & 0.182 & 0.187 & 0.155 \\
\textbf{MuCoCO}~\cite{kumar2021controlled}   & 0.326 & \textbf{0.325} & 0.278 & 0.218 & 0.256 & 0.244 & 0.221 & 0.318 & 0.242 & 0.254 \\
\textbf{SSD-LM}~\cite{han2022ssd}    & 0.247 & 0.246 & 0.233 & \textbf{0.283} & 0.318 & 0.298 & 0.261 & \textbf{0.321} & \textbf{0.279} & \textbf{0.274}  \\
\textbf{Prefix Tuning}~\cite{li2021prefix}  & 0.231 & 0.217 & 0.230 & - & 0.224 & 0.255 & - & 0.268 & 0.246 & - \\

\midrule[0pt]
\multicolumn{11}{c}{\textsc{\textbf{BLEU} - Our Approaches}} \\
\midrule[0pt]

\textbf{\textsc{Cev-LM} ($\mathcal{N}$-only)} & \textbf{0.340} & \textbf{0.327} & 0.305 & 0.246 & \textbf{0.329} & 0.268 & \textbf{0.287} & 0.249 & 0.268 & 0.248 \\
\textbf{\textsc{Cev-LM}} & 0.326 & 0.313 & 0.295 & 0.273 & 0.304 & 0.265 & 0.252 & 0.290 & 0.276 & 0.162 \\

\bottomrule
\end{tabular}}
\label{tab:bleu_bert}
\vspace{10pt}
\end{table*}

In the former part of~\Cref{tab:bleu_bert}, we report the BERT Scores of all approaches across various target deltas for each nonstandard control condition. We observe that \textsc{Cev-LM} consistently outperforms the other approaches while performing about on par with the edit-then-prototype baseline. This demonstrates that our approach preserves semantic similarity while significantly changing the speed, volume, or circuitousness of the text (also seen in~\Cref{tab:eval}). As the target delta increases, the BERT Score of our approach tends to decrease. We explain this phenomenon further in~\Cref{subsec:data_dist}. We include the scores for formality and toxicity in~\Cref{tab:other_bleu_bert}.

\subsection{Lexical Similarity}
\label{subsec:lexical_sim}

In the latter part of~\Cref{tab:bleu_bert}, we report the BLEU Scores of all approaches across various target deltas for each nonstandard control condition. We generally observe similar trends for our approaches in that BLEU score decreases as target delta increases, although to a greater scale. Since BLEU score measures lexical similarity, it is more sensitive to changes in wording, leading to a larger spread of scores. We also find that unlike before, \textsc{Cev-LM} does not clearly outperform the other approaches, which may imply that it presents semantically similar content while changing the wording. We include the scores for formality and toxicity in~\Cref{tab:other_bleu_bert}.

\subsection{Tolerance Tuning}
\label{subsec:tolerance}

We measure the impact of $\epsilon$ on $\Delta$ in~\Cref{tab:eval-tol} (see~\Cref{appendix:tolerance}), testing $\epsilon = \{0.05, 0.1, 0.2\}$. When tolerance is too low, the approach overfits from the lack of training data, leading to smaller percent errors and poor similarity metrics. Including controlled perturbations to edit vectors improves similarity metrics at the cost of $\Delta$, indicating that the approach may help combat overfitting.  The effect of controlled edit vector perturbation is inconsistent across tolerance values and attributes, so we use $\epsilon = 0.1$ to provide an effective balance of control over a certain $\Delta$ and enough data for robust training. More details can be found in~\Cref{appendix:tolerance}.

% It is likely that the perturbation benefits from having $\Delta$-constrained pairs but we leave it to future work to investigate this behavior.

\subsection{Training Delta Distribution}
\label{subsec:data_dist}

\begin{figure*}[!htbp]
  \centering
  \caption{Histogram of delta values (\ie $s(x) - s(x')$) within the Yelp Restaurant Review Corpus. The x-axis represents the difference in speed within the pairs of our created neighborhood, $\mathcal{N}(x)$, without any constraint on speed. The y-axis counts the number of pairs exhibiting the given delta in log-scale.}
  \subfigure[]{\includegraphics[width=0.32\textwidth]{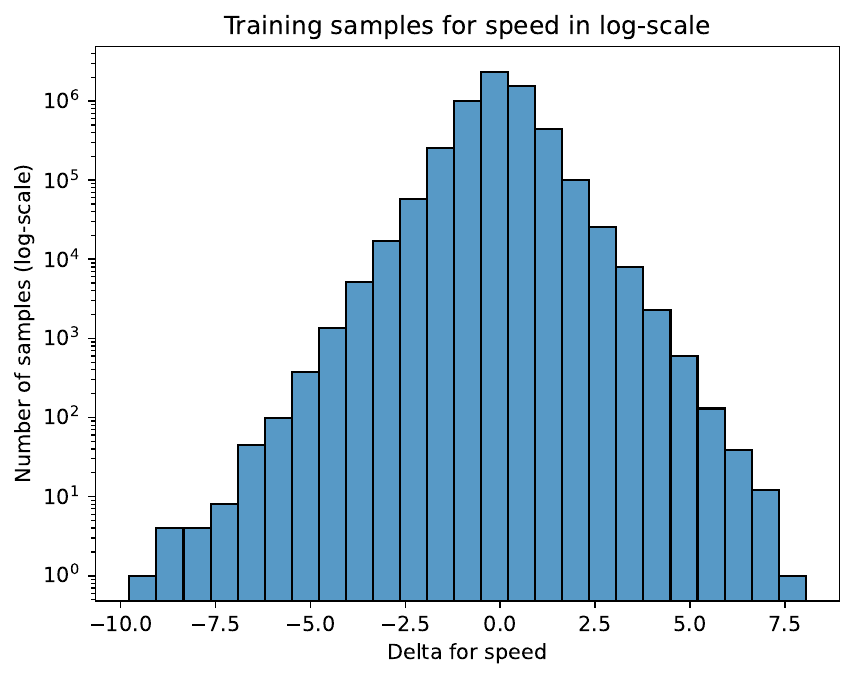}}
  \subfigure[]{\includegraphics[width=0.32\textwidth]{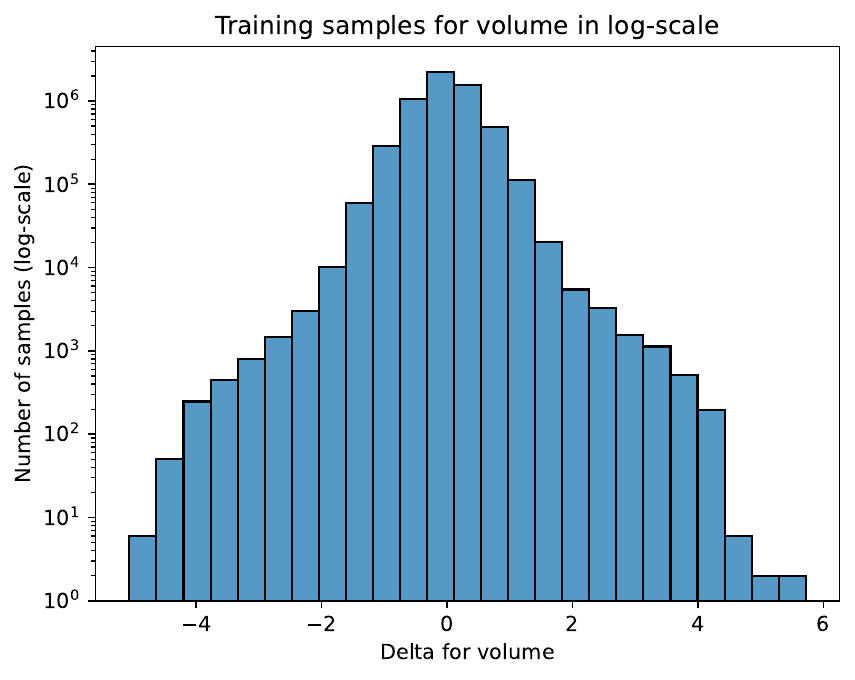}}
  \subfigure[]{\includegraphics[width=0.32\textwidth]{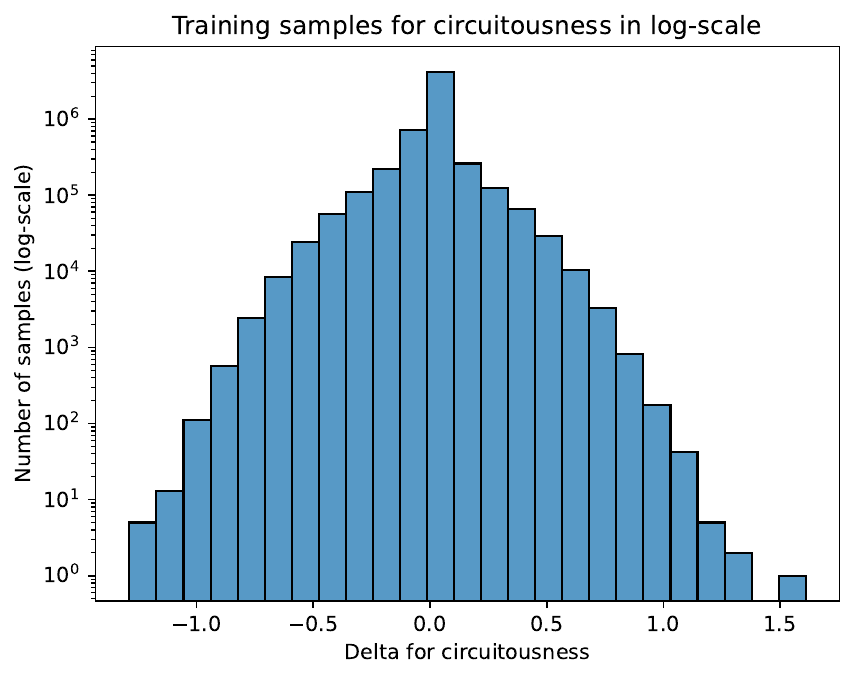}}
  \label{fig:feature_distr}
  \vspace{-10pt}
\end{figure*}

We analyze the distribution of $\Delta$ in $\mathcal{N}(x)$ in~\Cref{fig:feature_distr}. The distribution is centered at 0 and denser at smaller magnitudes of $\Delta$, implying a lack of training data for larger shifts in the target attribute. This explains the general trend of increasing MAE and decreasing BLEU/BERT score as the target delta increases, seen in~\Cref{tab:eval}.

\begin{figure}[!htbp]
  \centering
  \caption{The number of samples used for training versus percent error. The attribute is denoted by the shape, target delta by the color, and the border indicates that perturbation was used. Despite access to significantly fewer samples, low-resource models exhibit similar amounts of control to high-resource models.}
  \scalebox{0.7}{
  \includegraphics[width=0.68\textwidth]{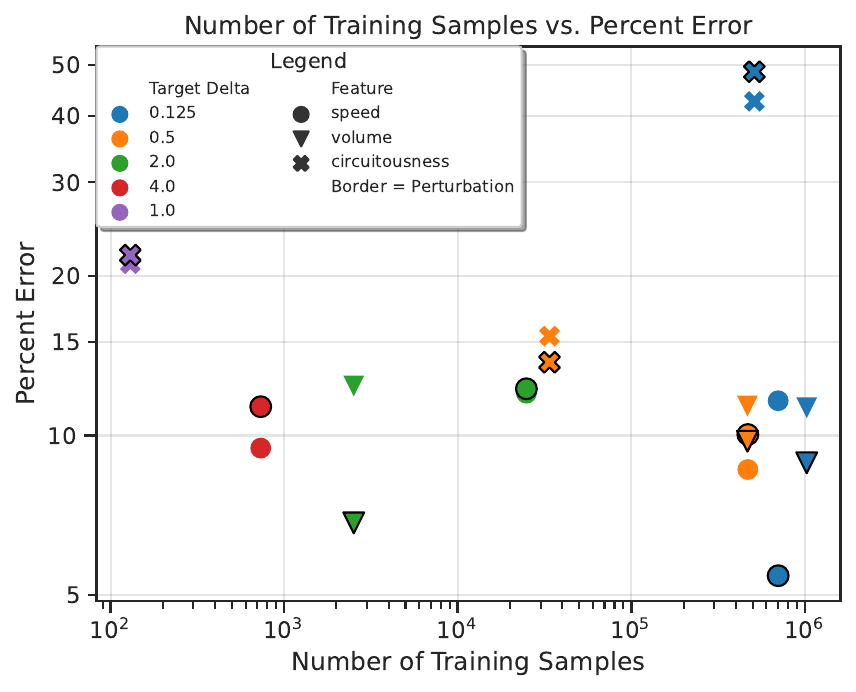}
  }
  \label{fig:samples_vs_error}
  \vspace{-20pt}
\end{figure}

In~\Cref{fig:samples_vs_error}, we record the performance against the number of training samples, finding that despite fewer samples, control through low-resource training is just as successful as through high-resource training. While circuitousness is the worst-performing attribute, likely due to the complexity of capturing the shortest path-based computation, it surprisingly does worst in a high-resource setting. It is possible that the change in attribute was too small to capture, even with a high number of samples. We see some success with decreasing the number of samples by a few orders of magnitude while preserving performance, across all attributes but leave extensive investigation to future work.

\subsection{Qualitative Results}
\label{subsec:qualitative}

\Cref{tab:ne_speed_example,tab:ne_volume_example,tab:ne_circuitousness_example} show the generations of the \textsc{Edit-then-Prototype} Baseline, GPT-3, SSD-LM, and each of our methods (\ie \textsc{Cev-LM ($\mathcal{N}$-only)} and \textsc{Cev-LM}) with $\epsilon=0.1$ when given a target delta of $0.5$ for speed, volume, and circuitousness, respectively. 
In~\Cref{tab:ne_speed_example}, we observe that our approaches generations tend to convey the same information in a shorter span, indicating an increase in speed. GPT-3 is generally not consistent, and SSD-LM tends to stray off-topic. In~\Cref{tab:ne_volume_example}, we can see that our approach tends to add information/verbosity, indicating an increase in volume. Here, GPT-3 shows little to no change in generation, and again, SSD-LM tends to stray off-topic, hallucinating information. Lastly, in~\Cref{tab:ne_circuitousness_example}, our approach leads to more indirect descriptions. While the text is more verbose, like volume, it repeats information/words, a key facet of circuitousness. GPT-3 hardly changes the input, and SSD-LM hallucinates some information. 

In addition to qualitative observations comparing variations of our approach,~\Cref{tab:eval} in~\Cref{appendix:perturb} shows the results of training without and with controlled edit vector perturbation, respectively, in comparison to a baseline edit-then-prototype model to show that our approach significantly changes the behavior of the edit-then-prototype model. We also include~\Cref{appendix:retrieval} to show qualitative examples of the retrieved prototypes.

  \begin{table*}[!htbp]
  \small
  \centering
  \caption{Generations of the baseline \cite{guu2018generating} ($\textbf{\textsc{Edit-then-Prototype}}$) as well as the benchmarks and \textsc{Cev-LM} architectures for a change in $\textbf{speed}$ with a target delta of $0.5$. We use a tolerance $\epsilon = 0.1$ for all of our approaches, as it empirically provided the best results in~\Cref{subsec:tolerance}. The models are fed the input (\ie \textbf{Original}) and generate by applying an ``edit vector'' to the latent representation of the input sentence.}
  \vspace{-5px}
  \label{tab:ne_speed_example} 
  \begin{tabularx}{\linewidth}{cl}
   \toprule[1.5pt]
  & \textsc{Model} \& \textsc{Generated Text} - \textbf{\textsc{Speed}}\\
  \midrule[0.75pt]
  % \textsc{Example 1:}\\
  & \textbf{Original}: He went above and beyond in providing us excellent customer service and was extremely \\ & \hspace{36pt} courteous friendly and kind.  \\
  & \textbf{\textsc{Edit-then-Prototype}}: He went above and beyond in providing us with amazing customer service and \\ & \hspace{36pt} was extremely courteous friendly and kind.  \\
  \textsc{Example 1} & \textbf{GPT-3}: He was a great customer service provider. He was friendly and kind, \\
  & \textbf{SSD-LM}: I was pleased that this was in line with our expectations. Suggesting the right\\
  & \textbf{\textsc{Cev-LM} ($\mathcal{N}$-only)}: Always amazing customer service and very knowlegable staff. \\
  & \textbf{\textsc{Cev-LM}}: He was knowledgable, courteous, and went provided excellent customer service.\\
  % \addlinespace[0.5em]
  \midrule[0.75pt]
  % \textsc{Example 2:}\\
  & \textbf{Original}: The staff is very professional and friendly \& environment is clean.  \\
  & \textbf{\textsc{Edit-then-Prototype}}: The staff is very professional and likable \& the environment is clean. \\
  & \textbf{GPT-3}: The staff was very professional, but not too friendly. The environment was clean, \\
  \textsc{Example 2} & \textbf{SSD-LM}: This is a great experience! Reviews the best experience in the world. \\
  & \textbf{\textsc{Cev-LM} ($\mathcal{N}$-only)}: The friendly staff is very professional \& environment is clean. \\
  & \textbf{\textsc{Cev-LM}}: Friendly staff and clean environment.\\
  \bottomrule[1.5pt]\\
  \end{tabularx}
  \vspace{-10px}
  \end{table*}

  \begin{table*}[!htbp]
  \small
  \vspace{-5pt}
  \centering
  \caption{Generations of the benchmarks and \textsc{Cev-LM} architectures for a change in $\textbf{volume}$ with a target delta of $0.5$. We use a tolerance $\epsilon = 0.1$ for all of our approaches. The models are fed the input (\ie \textbf{Original}) and generate by applying an ``edit vector'' to the latent representation of the input sentence.}
  \vspace{-5px}
  \label{tab:ne_volume_example} 
  \begin{tabularx}{\linewidth}{cl}
   \toprule[1.5pt]
  & \textsc{Model} \& \textsc{Generated Text} - \textbf{\textsc{Volume}}\\
  \midrule[0.75pt]
  % \textsc{Example 1:}\\
  & \textbf{Original}: Prices were very reasonable for the quality and quantity served.  \\
  & \textbf{\textsc{Edit-then-Prototype}}: Prices were very reasonable for the quality and quantity of food there. \\
  & \textbf{GPT-3}: The food was good, but the prices were a bit high for what \\
  \textsc{Example 1} & \textbf{SSD-LM}: The manager seemed very pleased with the quantity and the good service. The manager was \\ & \hspace{37pt} impressed by the price and quality. \\
  & \textbf{\textsc{Cev-LM} ($\mathcal{N}$-only)}: The prices were quite reasonable for the quality and quantity of food that was presented. \\
  & \textbf{\textsc{Cev-LM}}: And prices were surprisingly reasonable for the quality and quantity of food that was presented. \\
  % \addlinespace[0.5em]
  \midrule[0.75pt]
  % \textsc{Example 2:}\\
  & \textbf{Original}: I like this place--definitely going back. \\
  & \textbf{\textsc{Edit-then-Prototype}}: I like going to this place for lunch. \\
  & \textbf{GPT-3}: I really like this place. I'm definitely going back. \\
  \textsc{Example 2} & \textbf{SSD-LM}: The service was quite good. The food was available and delicious for two hours. \\
  & \textbf{\textsc{Cev-LM} ($\mathcal{N}$-only)}: Overall,everything was great and I'll be coming again next time. \\
  & \textbf{\textsc{Cev-LM}}: Overall,had a great time and I'll definitely be back. \\
  \bottomrule[1.5pt]\\
  \end{tabularx}
  \vspace{-10px}
  \end{table*}

  \begin{table*}[!htbp]
  \small
  \vspace{-5pt}
  \centering
  \caption{Generations of the benchmarks and \textsc{Cev-LM} architectures for a change in $\textbf{circuitousness}$ with a target delta of $0.5$. We use a tolerance $\epsilon = 0.1$ for all of our approaches. The models are fed the input (\ie \textbf{Original}) and generate by applying an ``edit vector'' to the latent representation of the input sentence.}
  \vspace{-5px}
  \label{tab:ne_circuitousness_example} 
  \begin{tabularx}{\linewidth}{cl}
   \toprule[1.5pt]
  & \textsc{Model} \& \textsc{Generated Text} - \textbf{\textsc{Circuitousness}}\\
  \midrule[0.75pt]
  % \textsc{Example 1:}\\
  & \textbf{Original}: I've only tried their Thai food,so far,and it is very good. \\
  & \textbf{\textsc{Edit-then-Prototype}}: I've only tried their Thai food,so far,and it is very good. \\
  & \textbf{GPT-3}: I've only tried their Thai food, so far, and it was really \\
  \textsc{Example 1} & \textbf{SSD-LM}: They are very good in preparing food. However, if the food is really not good, then the chicken \\ & \hspace{37pt} you can eat. \\
  & \textbf{\textsc{Cev-LM} ($\mathcal{N}$-only)}: I've decided that their Thai food is really good...their Japanese food,not so much. \\
  & \textbf{\textsc{Cev-LM}}: I've decided that their Thai food is really good...their Japanese food,not so much. \\
  % \addlinespace[0.5em]
  \midrule[0.75pt]
  % \textsc{Example 2:}\\
  & \textbf{Original}: I'd give the decor 4 stars and the food 3 stars. \\
  & \textbf{\textsc{Edit-then-Prototype}}: I'd give the service 2 stars and the food 3 stars. \\
  & \textbf{GPT-3}: I'd give the decor 3 stars and the food 4 stars. \\
  \textsc{Example 2} & \textbf{SSD-LM}: The food was not well-priced and expensive, but very well-made, and I was very pleased with it. \\
  & \textbf{\textsc{Cev-LM} ($\mathcal{N}$-only)}: 3 stars for the food and 2 stars for the prices equals 2.5 stars for me. \\
  & \textbf{\textsc{Cev-LM}}: I'd give 4 stars for the food and 3 stars for the service,3 stars for the decor. \\
  \bottomrule[1.5pt]\\
  \end{tabularx}
  \vspace{-15px}
  \end{table*}

\section{Conclusion}
\label{sec:conclusion}

In this work, we present \textsc{Cev-LM}, an inexpensive, semi-autoregressive language model that uses constrained edit vectors for controllable text generation. We compose an extensive set of controllable text generation benchmarks, and through quantitative and qualitative evaluations, we show that our approach leads to significantly substantially more control over nonstandard control conditions (\eg speed, volume, circuitousness) while preserving semantic meaning. 

Steerable natural language generation remains an open challenge, and we plan to continue improving our work in various directions, such as using a weighted mixture of \textsc{Cev-LM} models to capture all potential target deltas and replacing pieces of our architecture with larger language models. Ultimately, we hope to apply these models to subjective traits like memorability and persuasiveness, which are compositions of many smaller constraints (\eg conciseness, readability, etc.).

% an adaptation of the prototype-then-edit model proposed by \citet{guu2018generating} for controllable text generation. 

\newpage

\section*{Limitations}
While our approach shows substantial control over target attributes, adjusting our formulation to a wide range of controls may be tricky. Neighborhood creation can easily be adapted for any control but severely restricts the amount of training data. Perturbation works well with constraints defined with word embeddings due to how the edit vector is constructed, but it may struggle with other controls. While our approach works in scenarios with sparse training data, the quality of the training data still plays a significant role due to the prototyping step. A higher quality dataset with a large variety of sentences will lead to more diverse and well-suited generations.

Our approach lies in the retraining/refactoring category of controllable text generation models \cite{Zhang2022ASO}. Thus, it requires separate training for every attribute and target delta, which can be expensive as the model is scaled up. While most of our models can be trained for 100,000 steps on a single V100 in under a day\footnote{All of our experiments used roughly 1200 GPU hours, including training of baselines.}, we hypothesize that we can use a weighted composition of trained models to achieve any target delta. We leave it to future work to achieve such a framework. 

Lastly, in this study, we focus on the numerical control of non-standard control conditions. However, for humans, it is naturally better to quantize the values (e.g., higher vs. slightly higher speed). We choose numerical over categorical controls because fine-grained, numerical control over these features is less explored and more challenging. While it is difficult to go from categorical to numerical control, it is much easier to do the opposite direction - the main challenge is setting the boundaries of the categories.

% Ultimately, we hope to apply these models to course-grained traits like persuasiveness with many smaller constraints (\eg speed). However, naively using all constraints together would limit the size of the training data and pollute the edit vector perturbation. We leave it to future work to find innovative ways to combine 

\section*{Ethics Statement}
By nature of being trained on data from the internet and because large language models tend to memorize patterns without understanding the language or implications, our approach is susceptible to generating incorrect~\cite{zellers2019defending, maynez2020faithfulness, pagnoni2021understanding} or biased information as well as toxic language~\cite{wallace2019universal, gehman-etal-2020-realtoxicityprompts, sheng2021societal}. Although most studies have been conducted on autoregressive frameworks~\cite{bender2021dangers}, \textsc{Cev-LM} is still prone to such problems and future research is necessary to mitigate these issues. However, our framework attempts to achieve controllable outcomes, and future work can experiment with utilizing controllability to address the aforementioned challenges~\cite{liu-etal-2021-dexperts, han2022ssd}. Conversely, controllability can be utilized for malicious use cases, and we should ensure that future work continues to defend against such use cases by ensuring released data and models are protected against harmful/de-anonymized content.

\section*{Acknowledgements}

This work used the Extreme Science and Engineering Discovery Environment (XSEDE) Expanse GPU cluster, which is supported by National Science Foundation grant number ACI-1548562~\cite{xsede}. This work was also supported by the National Center for Supercomputing Application's Nano and Delta clusters.

\bibliography{anthology,custom}
\bibliographystyle{acl_natbib}

\appendix

\section{GPT-3 Baseline}
\label{appendix:gpt}

We construct few-shot prompts for controllable natural language generation with GPT-3. For all experiments, we use the ``davinci'' model, a temperature $\tau=0.7$, and test 1000 samples per attribute. The prompts are constructed in three parts:

\begin{itemize}
    \item \textbf{Attribute Description}: We begin all prompts by describing the attribute being used. In the case of speed, volume, and circuitousness, we find that providing both an intuitive explanation as well as a more mathematical definition leads to better results. For speed, we use the following:

    \begin{displayquote}
        Speed is a measure of how quickly content moves in a given text and is calculated as the distance traveled by consecutive windows of text. More specifically, we break the text into three-word chunks, compute the word embeddings of every chunk, and compute speed as the average distance between consecutive chunks.
    \end{displayquote}

    For volume, we use the following:
    
    \begin{displayquote}
        Volume captures the amount of information covered in a piece of text. We break the text into three-word chunks, compute the word embeddings of every chunk, and compute volume as the size of the minimum volume ellipsoid that contains all chunk embeddings.
    \end{displayquote}

    For circuitousness, we use the following:

    \begin{displayquote}  
        Circuitousness measures how indirectly content is covered. We break the text into three-word chunks, compute the word embeddings of every chunk, and compute circuitousness as the sum of distances between consecutive chunks divided by the length of the shortest path. The length of the shortest path is obtained by solving the traveling salesman problem.
    \end{displayquote}

    \item \textbf{Examples}: We continue the prompt with a set of $n$ examples to demonstrate how the attribute changes between sentences. These examples are randomly sampled from our training data, and one is shown below for speed:

    \begin{displayquote}  
        Sentence 1: PROS: Italian hoagie was delicious.  Friendly counter employee. The restaurant was clean and neat. 
                
        Generate a sentence such that the difference in speed between sentence two and sentence one is -0.3795
        
        Sentence 2: Great neighborhood Italian restaurant, especially in a neighborhood
        overrun by Italian restaurants. Love their white pizza. Small place, but very clean
        with super friendly staff.
    \end{displayquote}

    We use $n=3$ in our experiments because of the cost per token.
    
    \item \textbf{Prompt}: Lastly, we include the prompt, which uses three inputs: the original text, the attribute, and the target delta. The prompt is as follows:

    \begin{displayquote}  
        Sentence 1: \textsc{TEXT}
        
        Generate a sentence such that the difference in \textsc{ATTRIBUTE} between sentence two and sentence one is \textsc{TARGET DELTA}

        Sentence 2: 
    \end{displayquote}
    
\end{itemize}

\section{Attribute Classifier/Regressor}
\label{appendix:crt}

In this section, we provide further information about the training of the classifier and regressor used in our baseline models (\eg SSD-LM, MuCoCO, etc.). We train $\text{roberta-base}$\footnote{Available at \url{huggingface.co}. We also experimented with $\text{gpt-2}$ and $\text{bart-base}$ but found RoBERTa to be the most performant.}\cite{liu2019roberta} for 5 epochs or until training saturates (using an early stopping module), using an Adam optimizer with a learning rate of $5e-5$ and a batch size of $128$\footnote{We determine these parameters based on a simple grid search}. All other parameters are set based on the defaults provided by Huggingface~\cite{wolf2019}. These models contain roughly 125M parameters with 12 layers, 12 attention heads, and a hidden dimension of 768. We train each on roughly 2.6M samples from the Yelp dataset~\cite{yelp2017}.

For both MuCoCo and Prefix-Tuning, we utilize the regressor. We present the mean absolute error (MAE) in~\Cref{tab:clf_reg_eval} under $\mathcal{D}\textsc{-MAE}$ and normalized mean absolute error under $\mathcal{D}\textsc{-NMAE}$. To compute NMAE, we simply divide the MAE by the range of possible values. Generally, the MAEs are relatively small compared to the scale of $\Delta$, indicating a strong regressor. This is also reflected by the NMAEs. We try using a classifier with binned values for the attributes but find that the regressor performs better.

For SSD-LM, we trained a classifier to predict a binned difference in attributes such that all bins contain an equal number of training samples. We try adapting the formulation to work with a regressor but find that the classifier is substantially stronger. We record both F1-score and the mean absolute error (MAE) between classes in~\Cref{tab:clf_reg_eval} under $\mathcal{C}\textsc{-F1}$ and $\mathcal{C}\textsc{-MAE}$, respectively. We include both metrics to evaluate how well the model performs and roughly how incorrect predictions are. We observe that in most cases, the class labels are only at most off by one due to the low MAEs, indicating a strong classifier.

\section{\textsc{Cev-LM} Details}
\label{appendix:cevlm_params}

In this section, we provide more information about the training of \textsc{Cev-LM}. We encourage readers to reference our implementation for more details. All code and data are accessible at \url{https://drive.google.com/file/d/10rwCLJ96eNP5LS_1sG-flWvXD9X4pbjO}. Aside from the following parameters, our setting is identical to that of~\cite{guu2018generating}.

\subsection{Training}
In this paper, we train \textsc{Cev-LM} with a learning rate of $1e-3$ and batch size of $128$ for a maximum of $400,000$ iterations or a maximum wall time of 24 hours, whichever came first. The volume of data used for training depends on the definition of the constrained neighborhood, but generally, the most we use for a single model is roughly 1M samples.

\subsection{Model}
\textsc{Cev-LM} is extremely configurable, allowing you to switch out the encoder/decoder architecture and change aspects of the model, including the edit vector dimension, hidden dimension, and number of layers, among other features. In this paper, we use a simple attention mechanism~\cite{vaswani2017attention}, but future works can easily use larger language models in place of this mechanism to improve performance. We use an edit vector dimension of 256, a hidden dimension (for the encoder/decoder) of 256, 300-dimensional GLoVE~\cite{pennington2014glove} vectors. For the neural editor, we use 6 encoder and 6 decoder layers. For the inverse neural editor, we use 6 attention layers. In total, our checkpoint consists of roughly 76 million parameters (304MB).

\section{Prototyping Qualitative Analysis}
\label{appendix:retrieval}

In this section, we show some qualitative examples of the prototypes from our approach.~\Cref{tab:retrieval_example_speed,tab:retrieval_example_volume,tab:retrieval_example_circuitousness} include the input as well as the retrieved sample and the generated text. In many cases, we observe that the retrieved example demonstrates a strong change in feature, and \textsc{Cev-LM} corrects the strength of the change to ensure it is closer to the target delta.

  \begin{table}[htb]
  \small
  \centering
  \caption{Examples of an input, retrieved, and edited sentence for the model trained for a change in $\textbf{speed}$ with a target delta of $0.5$.  We use a tolerance $\epsilon = 0.1$ for our approach, as it empirically provided the best results in~\Cref{subsec:tolerance}. The models are fed the input (\ie \textbf{Original}) and generate by applying an ``edit vector'' to the latent representation of the input sentence.}
  \label{tab:retrieval_example_speed} 
  \begin{tabularx}{\linewidth}{@{}>{\raggedright\arraybackslash}X@{}}
   \toprule[1.5pt]
  \textsc{Retrieved} \& \textsc{Generated Text} - \textbf{\textsc{Speed}}\\
  \midrule[0.75pt]
  \textsc{Example 1:}\\
  \textbf{Input}: I will not return, terrible customer service. \\
  \textbf{Prototype}: Poorest customer service skills. \\
  \textbf{\textsc{Cev-LM}}: Terrible , terrible customer service. \\
  \addlinespace[0.5em]
  \textsc{Example 2:}\\
  \textbf{Input}: The food in the restaurant can be a little pricey, but it's good and you get a lot of it.  \\
  \textbf{Prototype}:  The food in the restaurant is a bit pricey, but it's good. \\
  \textbf{\textsc{Cev-LM}}: The food is good, but it's pricey. \\
  \bottomrule[1.5pt]\\
  \end{tabularx}
  \vspace{-10px}
  \end{table}

  \begin{table}[htb]
  \small
  \centering
  \caption{Examples of an input, retrieved, and edited sentence for the model trained for a change in $\textbf{volume}$ with a target delta of $0.5$.  We use a tolerance $\epsilon = 0.1$ for our approach, as it empirically provided the best results in~\Cref{subsec:tolerance}. The models are fed the input (\ie \textbf{Original}) and generate by applying an ``edit vector'' to the latent representation of the input sentence.}
  \label{tab:retrieval_example_volume} 
  \begin{tabularx}{\linewidth}{@{}>{\raggedright\arraybackslash}X@{}}
   \toprule[1.5pt]
  \textsc{Retrieved} \& \textsc{Generated Text} - \textbf{\textsc{Volume}}\\
  \midrule[0.75pt]
  \textsc{Example 1:}\\
  \textbf{Input}: Overall, this was a positive experience.  \\
  \textbf{Prototype}: Overall, we had a positive experience and the food was good. \\
  \textbf{\textsc{Cev-LM}}: Overall, a very positive experience - I'll definitely be back. \\
  \addlinespace[0.5em]
  \textsc{Example 2:}\\
  \textbf{Input}: The menu had lots of options. \\
  \textbf{Prototype}: The menu leaves you with lots of options that you can customize. \\
  \textbf{\textsc{Cev-LM}}: The menu here has lots of options that we want to try. \\
  \bottomrule[1.5pt]\\
  \end{tabularx}
  \vspace{-10px}
  \end{table}

    \begin{table}[htb]
  \small
  \centering
  \caption{Examples of an input, retrieved, and edited sentence for the model trained for a change in $\textbf{circuitousness}$ with a target delta of $0.5$.  We use a tolerance $\epsilon = 0.1$ for our approach, as it empirically provided the best results in~\Cref{subsec:tolerance}. The models are fed the input (\ie \textbf{Original}) and generate by applying an ``edit vector'' to the latent representation of the input sentence.}
  \label{tab:retrieval_example_circuitousness} 
  \begin{tabularx}{\linewidth}{@{}>{\raggedright\arraybackslash}X@{}}
   \toprule[1.5pt]
  \textsc{Retrieved} \& \textsc{Generated Text} - \textbf{\textsc{Circuitousness}}\\
  \midrule[0.75pt]
  \textsc{Example 1:}\\
  \textbf{Input}: This is my favorite Szechuan restaurant in town. \\
  \textbf{Prototype}:  This is my favorite Szechuan restaurant, and probably my favorite Szechuan restaurant ever.  \\
  \textbf{\textsc{Cev-LM}}: This is my favorite Szechuan restaurant in town and probably in the world.\\
  \addlinespace[0.5em]
  \textsc{Example 2:}\\
  \textbf{Input}: The menu has a little bit of everything.  \\
  \textbf{Prototype}: The menu has a little bit of everything that you could want. \\
  \textbf{\textsc{Cev-LM}}: The menu has a little bit of this and a little bit of that. \\
  \bottomrule[1.5pt]\\
  \end{tabularx}
  \vspace{-10px}
  \end{table}

\section{Similarity Scores on Toxicity \& Formality}
\label{appendix:sim_other}

In this section, we present the similarity scores of the baseline approach and our approach over formality and toxicity as control attributes in~\Cref{tab:other_bleu_bert}. We find that the scores demonstrate our generations stay on topic, further indicating the robustness of our approach on more standard control attributes.

\begin{table*}[h]
\centering
\tiny
\caption{BLEU and BERT (F1) Scores for speed, volume, and circuitousness across all approaches for different target deltas. The scores are averaged across three training runs (inference runs for GPT-3). We use a tolerance $\epsilon = 0.1$ for all of our approaches, as it empirically provided the best results in~\Cref{subsec:tolerance}.\vspace{-5pt}}
%\footnotesize
\noindent\setlength\tabcolsep{2.9pt}
\fontsize{4pt}{4pt}\selectfont
\resizebox{\textwidth}{!}{%
\begin{tabular}{@{}p{0.20\linewidth}@{\hspace{2pt}}
% K{\newfactor\linewidth}K{\newfactor\linewidth}K{\newfactor\linewidth}@{\hspace{5pt}}
% K{\newfactor\linewidth}K{\newfactor\linewidth}K{\newfactor\linewidth}@{\hspace{5pt}}
% % K{\newfactor\linewidth}K{\newfactor\linewidth}K{\newfactor\linewidth}@{\hspace{5pt}}
% K{\newfactor\linewidth}K{\newfactor\linewidth}K{\newfactor\linewidth}@{}} \\
ccc@{\hspace{5pt}}ccc@{\hspace{5pt}}ccc@{}}
\toprule
\textbf{Metric} & \multicolumn{3}{c}{\textbf{Toxicity}}  & \multicolumn{3}{c}{\textbf{Formality}}\\
\cmidrule{2-4} \cmidrule(lr){5-7}
%\midrule
{\textbf{Target Delta}} & \textbf{0.1} & \textbf{0.5} & \textbf{0.9} & \textbf{0.1} & \textbf{0.5} & \textbf{0.9} \\

\midrule
\multicolumn{7}{c}{\textsc{\textbf{BERTScore} - Benchmark Approaches}} \\
\midrule[0pt]

\textbf{GPT-3}~\cite{brown2020language}       & 0.857 & 0.869    & 0.866 & 0.851 & 0.851  & 0.862  \\
\textbf{MuCoCO}~\cite{kumar2021controlled}    & 0.763 & 0.771    & 0.774 & 0.763 & 0.759  & 0.760  \\
\textbf{SSD-LM}~\cite{han2022ssd}             & 0.769 & 0.767    & 0.769 & 0.763 & 0.760  & 0.747 \\
\textbf{Prefix Tuning}~\cite{li2021prefix}    & 0.833 & 0.827    & 0.823 & 0.843 & 0.836  & 0.834  \\ 

\midrule[0pt]
\multicolumn{7}{c}{\textsc{\textbf{BERTScore} - Our Approaches}} \\
\midrule[0pt]

\textbf{\textsc{Cev-LM}}                      & 0.848 & 0.827    & 0.837 & 0.842 & 0.842  & 0.845   \\

\midrule
\multicolumn{7}{c}{\textsc{\textbf{BLEU} - Benchmark Approaches}} \\
\midrule[0pt]

\textbf{GPT-3}~\cite{brown2020language}       & 0.231 & 0.250    & 0.291 & 0.219 & 0.273     & 0.269 \\
\textbf{MuCoCO}~\cite{kumar2021controlled}    & 0.305 & 0.268    & 0.293 & 0.257 & 0.276     & 0.219 \\
\textbf{SSD-LM}~\cite{han2022ssd}             & 0.294 & 0.343    & 0.346 & 0.314 & 0.325     & 0.326   \\
\textbf{Prefix Tuning}~\cite{li2021prefix}    & 0.246 & 0.269    & 0.275 & 0.231 & 0.217     & 0.194 \\

\midrule[0pt]
\multicolumn{7}{c}{\textsc{\textbf{BLEU} - Our Approaches}} \\
\midrule[0pt]

\textbf{\textsc{Cev-LM}}                      & 0.320 & 0.316    & 0.295 & 0.342 & 0.334     & 0.265 \\

\bottomrule
\end{tabular}}
\label{tab:other_bleu_bert}
% \vspace{-10pt}
\end{table*}

% \begin{table}[]
% \begin{tabular}{lcccccc}
% \hline
% BERTScore     &       & Toxicity &       &       & Formality &        \\
% Target Delta  & 0.1   & 0.5      & 0.9   & 0.1   & 0.5    & 0.9    \\ \hline
% GPT-3         & 85.65 & 86.89    & 86.58 & 85.08 & 85.12  & 86.20  \\
% MuCoCO        & 76.31 & 77.08    & 77.41 & 76.31 & 75.92  & 76.01  \\
% SSD-LM        & 76.86 & 76.66    & 76.87 & 76.27 & 75.96  & 0.7472 \\
% Prefix Tuning & 83.21 & 82.73    & 82.27 & 84.34 & 83.60  & 83.39  \\ \hline
% CEV-LM        & 84.81 & 82.70    & 83.69 & 84.18 & 84.19  & 84.47 
% \end{tabular}
% \end{table}

% \begin{table}[]
% \begin{tabular}{lcccccc}
% \hline
% BLEU          &       & Toxicity &       &       & Formality &       \\
% Target Delta  & 0.1   & 0.5      & 0.9   & 0.1   & 0.5       & 0.9   \\ \hline
% GPT-3         & 0.231 & 0.250    & 0.291 & 0.219 & 0.273     & 0.269 \\
% MuCoCO        & 0.305 & 0.268    & 0.293 & 0.257 & 0.276     & 0.219 \\
% SSD-LM        & 0.294 & 0.343    & 0.346 & 0.314 & 0.325     & 0.326 \\
% Prefix Tuning & 0.246 & 0.269    & 0.275 & 0.231 & 0.217     & 0.194 \\ \hline
% CEV-LM        & 0.320 & 0.316    & 0.295 & 0.342 & 0.334     & 0.265
% \end{tabular}
% \end{table}

\section{Tolerance Tuning}
\label{appendix:tolerance}

\begin{table*}[]
  \centering
  \small
  \caption{Evaluation metrics (BLEU \cite{Papineni2002bleu} and BERTScore \cite{zhang2019bertscore}) and strength of control on $\Delta$ for the trained models (ideally, $\Delta = 0.5$) for speed. The scores are averaged across three training runs with different seeds. We train a baseline edit-then-prototype model \cite{guu2018generating}, as well as \textsc{Cev-LM ($\mathcal{N}$-only)} and \textsc{Cev-LM} with different tolerances ($\epsilon$). We record both train and test BLEU to demonstrate overfitting with lower tolerances. }
  \label{tab:eval-tol}
  \begin{tabularx}{0.9\linewidth}{@{}>{\raggedright\arraybackslash}Xcccc@{}}
   \toprule[1.5pt]
  \textsc{Model} & \textsc{Delta} & \textsc{Train BLEU} & \textsc{Test BLEU} & \textsc{BERTScore} \\     
  \midrule[0.75pt]
  % \addlinespace[0.5em]
\textsc{Edit-then-Prototype}    & 0.0113 & \textbf{0.6691}     & \textbf{0.5679}    & 0.9327 \\
  \midrule[0.75pt]
\textsc{Cev-LM ($\mathcal{N}$-only)}: $\epsilon$ = 0.05 & 0.4559 & 0.8057     & 0.4266    & 0.9326 \\
\textsc{Cev-LM ($\mathcal{N}$-only)}: $\epsilon$ = 0.1  & 0.4558 & 0.7146     & \textbf{0.5747}    & 0.9340 \\
\textsc{Cev-LM ($\mathcal{N}$-only)}: $\epsilon$ = 0.2  & 0.4405 & 0.5994     & 0.5628    & 0.9355 \\
  \midrule[0.75pt]
% Perturbation            & 0.4433 & 0.5513     & 0.4531    & 0.9346 \\
\textsc{Cev-LM}: $\epsilon$ = 0.05  & 0.4279 & 0.5709     & 0.5218    & 0.9329 \\
\textsc{Cev-LM}: $\epsilon$ = 0.1   & 0.4455 & 0.6375     & 0.5400    & 0.9386 \\
\textsc{Cev-LM}: $\epsilon$ = 0.2   & \textbf{0.4596} & \textbf{0.6751}     & \textbf{0.5679}    & 0.9334 \\
  \bottomrule[1.5pt]\\
  \end{tabularx}
  \vspace{-10pt}
  \end{table*}

We measure the impact of $\epsilon$ on training in~\Cref{tab:eval-tol}. Too low of a tolerance value leads to overfitting, indicated by a closer $\Delta$ to the target and poor performance in the test-time similarity metrics. As mentioned before, controlled edit vector perturbations to edit vectors improves similarity metrics at the cost of $\Delta$, which implies that the approach helps to combat overfitting. At $\epsilon = 0.05$ and $\epsilon = 0.1$, we see that perturbation is generally not helpful, but at tolerance $\epsilon = 0.2$, the perturbation approach leads to a higher $\Delta$. Note that the BLEU scores are slightly different as n-grams are weighted differently in the code for the edit-then-prototype architecture \cite{guu2018generating} and in NLTK \cite{bird-loper-2004-nltk}.

\section{Controlled Edit Vector Perturbation}
\label{appendix:perturb}

In~\Cref{tab:eval}, we present the results of the neighborhood creation and neighborhood creation + perturbation approaches. The Baseline shows the out-of-the-box edit-then-prototype model, which has little impact on the target attribute and provides a rough baseline of the similarity metrics. Again, we find that as the target delta increases, the MAE increases and similarity scores decrease. This phenomenon is attributed to the data distribution and is expanded on in~\Cref{subsec:data_dist}. We observe that perturbation is sometimes helpful in decreasing MAE, especially in the case of volume. However, this behavior is inconsistent across speed and circuitousness and warrants further exploration. 

  \begin{table*}[]
  \centering
  \caption{Evaluation metrics (BLEU \cite{Papineni2002bleu} and BERTScore \cite{zhang2019bertscore}) and strength of control on $\Delta$ for the trained models. The scores are averaged across three training runs, and we omit variance due to negligible values. We train a baseline model \cite{guu2018generating} (Baseline) and multiple models across various target deltas for all nonstandard control conditions (\eg speed, volume, circuitousness) to show training has a significant impact on the achieved control.}
  \label{tab:eval}
\scalebox{0.7}{
\begin{tabularx}{1.36\linewidth}{cccccc|ccccc}

                & \multicolumn{5}{c}{\textbf{\textsc{Cev-LM ($\mathcal{N}$-only)}}} & \multicolumn{5}{c}{\textbf{\textsc{Cev-LM}}}\\
                \toprule[1.5pt]
               & \textsc{Target Delta} & \textsc{Delta} & \textsc{MAE} & \textsc{BLEU} & \textsc{BERT-F1} & \textsc{Target Delta} & \textsc{Delta} & \textsc{MAE} & \textsc{BLEU} & \textsc{BERT-F1} \\
\midrule[0.75pt]
               & Baseline  & 0.0468  & -      & 0.3185 & 0.9327 & Baseline  & 0.0468  & -      & 0.3185 & 0.9327 \\
               & 0.125     & 0.1105  & 0.0145 & 0.3399 & 0.9351 & 0.125     & 0.1189  & 0.0061 & 0.3261 & 0.9350 \\
Speed          & 0.5       & 0.4558  & 0.0442 & 0.3276 & 0.9340 & 0.5       & 0.4355  & 0.0645 & 0.3123 & 0.9386 \\
               & 2.0       & 1.7594  & 0.2406 & 0.3051 & 0.9291 & 2.0       & 1.7897  & 0.2103 & 0.2944 & 0.9281 \\
               & 4.0       & 3.6213  & 0.3787 & 0.2463 & 0.9188 & 4.0       & 3.4657  & 0.5343 & 0.2736 & 0.9230 \\
\midrule[0.75pt]
               & Baseline  & 0.0011  & -      & 0.3185 & 0.9327 & Baseline  & 0.0011  & -      & 0.3185 & 0.9327 \\
Volume         & 0.125     & 0.1106  & 0.012  & 0.3296 & 0.9380 & 0.125     & 0.1130  & 0.012  & 0.3038 & 0.9351 \\
               & 0.5       & 0.4415  & 0.0585 & 0.2682 & 0.9320 & 0.5       & 0.4535  & 0.0465 & 0.2653 & 0.9314 \\
               & 2.0       & 1.7521  & 0.2479 & 0.2869 & 0.9244 & 2.0       & 1.8466  & 0.1534 & 0.2518 & 0.9208 \\
\midrule[0.75pt]
               & Baseline  & -0.0022 & -      & 0.3185 & 0.9327 & Baseline  & -0.0022 & -      & 0.3185 & 0.9327 \\
Circuitousness & 0.125     & 0.0723  & 0.0527 & 0.2483 & 0.9271 & 0.125     & 0.0664  & 0.0586 & 0.2902 & 0.9306 \\
               & 0.5       & 0.4217  & 0.0783 & 0.2680 & 0.9109 & 0.5       & 0.4207  & 0.0793 & 0.2755 & 0.9089  \\
               & 1.0       & 0.7893  & 0.2107 & 0.2479 & 0.9082 & 1.0       & 1.0519  & 0.0519 & 0.1622 & 0.8354  \\
\bottomrule[1.5pt]\\ \\
% \midrule[0.75pt] \\
\end{tabularx}
}
\end{table*}

\end{document}